\documentclass[10pt,journal,compsoc]{IEEEtran}
\ifCLASSOPTIONcompsoc
  \usepackage[nocompress]{cite}
\else
  \usepackage{cite}
\fi
\ifCLASSINFOpdf
\else
\fi

\usepackage{graphicx}
\usepackage{subfigure}
\usepackage{url}
\usepackage{amsmath}
\usepackage{amssymb}
\usepackage{units}

\usepackage{algorithm}
\usepackage{algorithmic}
\begin{document}
%
\title{Joint Intermodal and Intramodal Label Transfers for Extremely Rare or Unseen Classes}
%
%
%
%

\author{Guo-Jun~Qi, \IEEEmembership{Member,~IEEE,}
        Wei~Liu,
        Charu~Aggarwal,~\IEEEmembership{Fellow,~IEEE,}
        and~Thomas~Huang,~\IEEEmembership{Life~Fellow,~IEEE}
\IEEEcompsocitemizethanks{
\IEEEcompsocthanksitem G.-J. Qi is with the Department
of Computer Science, University of Central Florida, Orlando,
FL, 32816.\protect\\
E-mail: guojun.qi@ucf.edu
\IEEEcompsocthanksitem W. Liu and C. Aggarwal are with the IBM T.J. Watson Research Center, Yorktown Heights, NY 10598.\protect\\
E-mail: \{weiliu, charu\}@us.ibm.com
\IEEEcompsocthanksitem T. Huang is with the Department of Electrical and Computer Engineering, University of Illinois, Urbana, IL, 61801.\protect\\
E-mail: huang@ifp.uiuc.edu
}
}

%
%

\markboth{This paper has been accepted by IEEE Transactions on Pattern Analysis and Machine Intelligence}%
{Qi \MakeLowercase{\textit{et al.}}: Joint Inter-and-Intra Modal Label Transfer from Texts to Images for Rare or Unseen Classes, Volume: PP Issue: 99}
%



\IEEEtitleabstractindextext{%
\begin{abstract}
In this paper, we present a label transfer model from texts
to images for image classification tasks. The problem of image classification is often much
more challenging than text classification.  On one hand, labeled text data is more widely available than the labeled images for
classification tasks.  On the other hand, text data tends to have
natural semantic interpretability, and
they are often more directly related to class labels. On the contrary, the image
features are not directly related to concepts inherent in
class labels. One of our goals in this paper is to develop a model for revealing the
functional relationships between text and image features as to
{\em directly transfer intermodal and intramodal labels} to annotate the images.
This is implemented by learning a transfer function as a bridge to propagate
the labels between two multimodal spaces. However, the intermodal label transfers could be undermined by  blindly transferring the labels of noisy texts to annotate images.  To mitigate this problem, we present an intramodal label transfer process, which complements the intermodal label transfer by
transferring the image labels instead when relevant text is absent from the source corpus.
In addition, we generalize the inter-modal label transfer to zero-shot learning scenario where there are only text examples available to label unseen classes of images without any positive image examples.
 We evaluate our algorithm
  on an image classification task and show
the effectiveness with respect to the other compared algorithms.
\end{abstract}

\begin{IEEEkeywords}
Multimodal analysis, intermodal and intramodal label transfers (I2LT), image classification, zero-shot learning
\end{IEEEkeywords}}

\maketitle

\IEEEdisplaynontitleabstractindextext

%
\IEEEpeerreviewmaketitle

\IEEEraisesectionheading{\section{Introduction}\label{sec:introduction}}

Label transfer between different modalities
 is a problem of using the training examples in one modality (e.g., texts) to enhance
 the training process for another modality (e.g., images) \cite{Dai:NIPS2008}\cite{Qi:WWW11}.  This has been studied
 before in different tasks involving the data of diverse modalities.
 One of the most important applications is content-based search and semantic indexing for
text documents and images. The text documents are much easier to label
 as compared to associated images on a webpage. Also, since classifiers naturally work better with features
that have semantic interpretability, text features are  inherently friendly
to the classification process in a way that is often a challenge for
visual representations of images. It makes it easier to
interpret and solve the classification problem in the text modality, while there is
a tremendous semantic gap between visual features and the concepts for images. In addition,
the challenges of image classification are particularly evident,
when the amount of training data available is limited. In such
cases, the image classification is further hampered by the
paucity of labels.


 In the case of images, it is desirable to obtain a feature
representation which relates more directly to semantic concepts; a
process which will improve the quality of classification.
Furthermore, this often has to be achieved with the use of only a
limited amount of labeled image data.  This naturally motivates an
approach for utilizing the labeled data in the text modality in order
to improve image classification. Hence, we implemented an {\em intermodal label transfer} process in which
 a transfer function is built to reveal the alignment between modalities so the labels
  can be transferred across different modalities \cite{Qi:WWW11}.
We showed that the transfer of the rich
label information from texts to images provides much more effective learning algorithms.

Although intermodal label transfer has shown promising result \cite{Dai:NIPS2008}\cite{Qi:WWW11}, however, we have observed that it might fail when the labels cannot be well aligned between modalities.  For example, the text labels of ``building" may refer to a large variety of building architectures in different documents, while a test image of ``building" often has a certain style of appearance.  It is risky to blindly transfer text labels no matter when the visual appearance does not match with any text descriptions from a source corpus.  This causes the ``negative tranfer" problem that refers to transferring of irrelevant information between different modalities \cite{rosenstein2005transfer}.
To prevent the negative transfer, we will present an intramodal label transfer process to complement the intermodal label transfer, which will take over the annotation of a test image by transferring image labels in absence of labeled relevant text documents.
As a result, this yields a joint Intermodal and Intramodal Label Transfer (I2LT) algorithm, which combines the advantages of both image labels and text labels in the context of a label transfer task.

Formally, we seek to develop a label transfer algorithm for jointly sharing
labels across and within different modalities \cite{Wu:ICML04} \cite{Raina:ICML07}
\cite{Dai:NIPS2008}.  Specifically, it is applied to the image classification problem in order to leverage the labels in
text corpora to annotate image corpora with scarce labels.  Such
algorithms typically transfer labeling information between heterogeneous
feature spaces \cite{Dai:NIPS2008}\cite{Zhu:AAAI10}\cite{Yang:ACL09}
instead of homogeneous feature spaces \cite{Wu:ICML04}.
Heterogeneous transfer learning is usually much more challenging due
to the unknown alignment across the distinct feature spaces. In
order to bridge across  two distinct feature spaces, the key
ingredient is a ``transfer function" which can explain the
alignment between text and image feature spaces through the use
of a feature transformation. This transformation is used for the
purpose of effective image classification and semantic indexing.
As discussed earlier,
it is achieved with the use of  co-occurrence data that is
often available in many practical settings.  For example,
in many real web and social media applications, it is  possible to
obtain many {\em co-occurrence pairs} between text and images \cite{qi2012exploring}; in web pages, the images are surrounded by text descriptions on the same web
page. Similarly, there is a tremendous amount of linkage between
text and images on the web, through comments in image sharing sites,
posts in a social networks, and other linked text and image corpora.
 It is reasonable to assume that the content of the text and the
images are highly correlated in both scenarios. This information provides a {\em
semantic bridge}, which can be exploited in order to learn the
alignment and label transfer between the different modalities.

In contrast to previous
work \cite{Dai:NIPS2008}\cite{Zhu:AAAI10}\cite{Yang:ACL09},
the label transfer proposed in this paper  can
establish the alignment between texts and images even
if {\bf the new test images do not have any surrounding text description}, or if the co-occurrence data
is independent of the labeled source texts.  This increases the
flexibility of the algorithm and makes it more widely applicable
in many practical applications.  Specifically, in order to perform the label transfer process, we create a new
{\em topic space}  into which both the text and images are mapped.
Both the occurrence set and training set
are used to learn the transfer function, which aligns
heterogeneous text and image spaces.  We also follow the {\em principle of
parsimony}, and encode as few topics as possible in order to
align between text and images for regularization. This principle
has a preference for the least complex model, as long as the text
and image alignment can be well explained by the learned
transfer function. After the transfer function is learned, the labels can
be propagated from  any labeled text corpus to any  new image by intermodal label propagation. While
labels from the images are also used for
improving accuracy, one characteristic of our transfer function is that it
is particularly robust in the presence of a very small number of
scarce training examples.

The remainder  of this paper is organized as follows.  In Section 2, we briefly review the related work.
Then we propose an intermodal label transfer process in Section 3 and show how the labels of
text corpus can be propagated to image corpus.  In section 4, a joint
Intermodal and Intramodal Label Transfer (I2LT) process is proposed, along with a transfer function in Section 5 that instantiates the joint model. In Section 6,
we present the objective problem along with a proximal gradient based algorithm for solving the optimization problem. We also present a zero-shot learning extension of the proposed algorithm to classify images of unseen classes in Section 7.
The experiment results are presented in section 8.  The conclusion and summary is presented
in Section 9.

\section{Related Work}

A variety of transfer learning methods have been proposed in prior pioneering
works, e.g., domain adaption \cite{lipami2014domainadaptation,duan2009domain,lieccv2014prvilegedinformation,Wu:ICML04,Raina:ICML06,Raina:ICML07}, cross-category information sharing \cite{Qi:CVPR11}, and heterogeneous transfer learning \cite{Qi:WWW11,Zhu:AAAI10,Dai:NIPS2008,gongijcv2013multiview}. In this paper, we concentrate on learning cross-modal correspondence and sharing the semantic information across different modalities.

Learning semantic correspondence from text to images can
 be seen as a transfer learning  problem that involves
heterogeneous data points across different feature spaces. For example,
\cite{Zhu:AAAI10} proposes \emph{heterogeneous transfer learning}, which uses both user tags and related
document text as auxiliary information to extract a new latent feature
representation for each image. However, it does not utilize the text labels to enrich the
semantic labels of images, which may restrict its performance when
the image labels are very scarce.   On the other
hand, translated learning \cite{Dai:NIPS2008} attempts to label the
target instances through a Markovian chain.  A translator is assumed
to be available between source and target data for correspondence.
However, given an arbitrary new image, such a correspondence is not
always directly available between any text and image instances.  In
this case, a generative model is used in the Markovian chain to
construct feature-feature co-occurrence.  This model is not reliable
when co-occurrence data is noisy and sparse. On the contrary,  we
explicitly learn a semantic transfer function, which directly
propagates semantic labels from text to images even if the semantic
correspondence is not available beforehand for a new image.
It avoids overfitting into the noisy and sparse co-occurrence data by
imposing the prior of fewest topics on semantic translation.

It is also worth noting that learning label transfer across heterogenous modalities is different from the conventional
\emph{heterogeneous
learning}, such as multi-kernel learning
\cite{Bach:ICML04} and co-training \cite{Blum:COLT98}.  In
heterogeneous learning, each instance must contain different views.
On the contrary, when translating text to images \cite{Qi:WWW11}, {\em it is not
required that an image has an associated text view.} This makes the
problem much more challenging. The correspondence between text and
images is established by the learned transfer function, and a single image
view of an input instance is enough to predict its label by a
label transfer process.

We also distinguish the proposed label transfer model from the other
latent models.  Previous latent methods, such as Latent Semantic
Analysis \cite{Landauer:DP98}, Probabilistic Latent Semantic
Analysis \cite{Hofmann:UAI99}, Latent Dirichlet Allocation
\cite{Blei:JMLR03} and Multimodal Latent Attributes \cite{fu2014learning}, are restricted to latent factor discovery from
the co-occurrence observations.  On the contrary, in this paper, the
goal is to establish semantic bridge so that the discriminative labeling information can be propagated between
the source and target spaces.  To the best of our knowledge, it is
one of the first algorithms to address such heterogeneous label
transfer problem \emph{via a parsimonious latent topic space}.  It is worth noting that even with \emph{unknown correspondence to
source instances}, it can still label the new instance by predicting
its correspondence based on the learned transfer function.

We also note that the proposed label transfer problem also differs from the problem of translating images and videos into sentences of natural languages \cite{venugopalan2014translating}\cite{ordonez2011im2text}.  Usually Recurrent Neural Networks (RNNs) \cite{hochreiter1997long} are used as a mathematical tool to map the visual feature vectors into words via a sequence of intermediate representation of long short-term memory cells \cite{venugopalan2014translating}.  The translation problem has been recognized as a very challenging task, since it requires the machine not only capable of reading the content of images and videos accurately, but also be able to translate the visual elements into sentences in a correct order with a satisfactory level of grammatical correctness.  In this paper, we do not aim to solve this challenging problem.  Instead we consider the label transfer from texts to images, where we do not need to compose the sentences.  Also, our goal differs from composing the description of the visual content in sentences.  Instead, we wish to utilize the abundant labeled text documents to improve the classification accuracy for the image classification tasks.

Although we focus on label transfer from texts to images, the model developed in this paper is equally applicable to the other label transfer tasks between different modalities.  For example, the previous work has demonstrated an application where the labels of English documents are transferred to annotate the Chinese documents \cite{Dai:NIPS2008}.  Similarly, the speech segments can be aligned by learning a transfer function by which the labels can be transferred across different languages to annotate the speeches.  The label transfer model can also be applied for audio-video recognition tasks \cite{moon2014multimodal,chen2014audio}.  Similar to the scenario set in this paper, a test audio will have no paralleled video and it must be aligned to the existing corpus of videos to enable intermodal label transfer.  But \cite{moon2014multimodal} explores a slightly different idea -- instead of aligning the test sample with the video corpus, they attempt to reconstruct the paralleled video through multimodal deep networks \cite{ngiam2011multimodal}.  This approach is indirect for label transfer and an independent classifier must be trained for audio-video recognition tasks.

In an earlier work \cite{gongijcv2013multiview}, Flickr images with tags have been used to learn several CCA variants for cross-modal retrieval task.  They incorporated a third view of supervised semantic information or unsupervised word clusters to bridge the cross-modal gap, along with the visual and text views.  On the contrary, an important byproduct of the proposed algorithm is a intermodal transfer function, which can measure the cross-modal relevance directly.  It is also learned with the supervised labeled image/text pairs.  In this spirit, our approach also involves a ``third view" of the labeled concepts.  However, our approach is motivated to annotate the labels of semantic concepts, rather than learning the cross-modal relevance directly.  This makes the proposed approach in a complimentary technical line to those CCA variants presented in \cite{gongijcv2013multiview}.

A more recent work \cite{niu2015exploiting} proposed to use privileged information to augment Support Vector Machines (SVMs).  Additional training bags were collected from textural descriptions of images, where positive bags contain the returned images containing relevant tags, while negative bags do not contain any images with relevant tags.  Then the problem with the training bags was formulated as multi-instance learning problem, and positive bags provide privileged information to augment the training of the classifiers with more positive instances of images.  Our approach differs from this method in directly transferring the text labels to reconstruct image labels, rather than training an image classifier.  However, both methods do not assume the availability of text information for testing images, making them applicable to label new images without text descriptions.

\section{Intermodal Label Transfer}
\begin{figure}[!t]
\vspace{5mm}
\begin{center}
\includegraphics[width=0.8\linewidth]{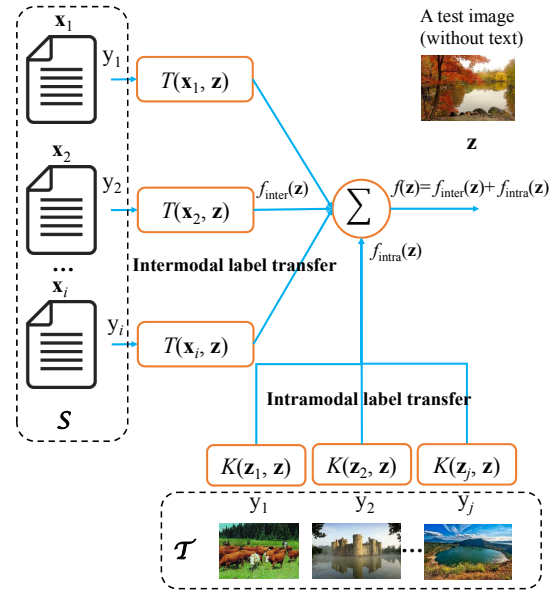}
\end{center}
 \caption{Illustration of semantic label propagation from text to images
 by the learned transfer function. On the left is the labeled text corpus $\mathcal S$, and at the bottom is the labeled image corpus $\mathcal T$.  The proposed I2LT transfers
 the labels from both corpora to annotate a test image at the top right corner. The output label is given by a discriminant
 function $f\left(\mathbf z\right)$.  Note that the test image is not associated with any text document. Hence, the transfer function $T$ is applied for the intermodal label transfer $f_{\rm inter}$ from source corpus $\mathcal S$, along with the intramodal label transfer $f_{\rm intra}$ from the image corpus $\mathcal T$.}\label{Fig:Fig01}
\end{figure}
In this section, we will introduce the notations and problem definitions for the label transfer process.
Let $\mathcal X\subset\mathbb R^{p}$ and $\mathcal Z\subset\mathbb R^{q}$ be the source and target
feature spaces, which have a dimensionality of $p$ and $q$
respectively. For the purpose of this paper, the source space
corresponds to the text modality, and the target space corresponds to
the image modality.  In the source (text) space, we have a set of
$n$ text documents in $\mathcal X$. Each text document is
represented by a feature vector $\mathbf x_i \in \mathcal X$. This text corpus, $\mathcal S = \left\{ {\left( {\mathbf x_i , y_i }\right)|1 \le i \le n } \right\}$, has already been annotated with class
labels, where $y_i \in
\left\{ { + 1, - 1} \right\}$ is the binary label for each document $\mathbf x_i$.
The binary
assumption is made to avoid notational clutter, and it can be straightforwardly extended
to encode multiple classes.

The images are represented by feature
vectors $\mathbf z$ in the target space $\mathcal Z$ . The task is to
relate the feature structure of the source (text) space to the
target space (image) space, so that the labeling information can be shared
between two spaces. The goal of the
transformation process is to provide a classifier for the target
(image) domain in the presence of scarce labeled data for the latter
domain.

In order to perform the label propagation from the text to the
image domain, we need a bridge, which relates the text and image
information. A key component which provides such bridging
information about the relationship between the text space $\mathcal X$ and image
feature space $\mathcal Z$ is a set of {\em co-occurrence pairs} ${\cal C} = \left\{
{\left( {\mathbf x}_k ,{\mathbf z}_k \right)}|{\mathbf x}_k\in \mathcal X, {\mathbf z}_k\in\mathcal Z, k=1,\cdots,l \right\}$. Such co-occurrence information
is abundant in the context of web and social network
data. In fact, it may often the case that the co-occurrence
information between text and images can be more readily obtained
than the class labels in the target (image) domain. For example, in
many web collections, the images may co-occur with the surrounding
text on the same web page. Similarly, in web and social networks, it
is common to have implicit and explicit links between text and
images. Such links can be viewed more generally as co-occurrence
data. This co-occurrence set provides the semantic bridge needed for
transfer learning.

  Besides the co-occurrence
set, we also have a small set $\mathcal T = \left\{
{\left( {\mathbf z_j ,y_j } \right)| \mathbf z_j\in\mathcal Z, 1 \le j \le m }
\right\}$ of labeled images. This is an auxiliary set of
training examples, and its size is usually much smaller than
that of the set of labeled source examples. In other words, we have
$m\ll n$.  As we will see, the auxiliary set is used in
order to enhance the accuracy of the transfer learning process.

One of the key intermediate steps during this process is the design
of a {\em transfer function}  between text and images. This
transfer function   serves as a conduit to  measure the alignment
 between text and image features. We will show that such a
conduit can  be used directly in order to  propagate the class
labels from text to images. The transfer function $T$ is defined jointly
on text space $\mathcal X$ and image space $\mathcal Z$ as
$T:\mathcal X \times \mathcal Z  \to \mathbb R $. It assigns a
real value to one pair of texts and image instances to weigh their
alignment. This value can be either positive or negative,
representing either positive or negative match. Given a new image
$\mathbf z$, its label is determined by an intermodal discriminant function as a
linear combination of the class labels in $\mathcal S$
weighted by the corresponding transfer functions
\begin{equation}\label{Eq:Eq01}
f_{\rm inter} \left( \mathbf z \right) = \sum\limits_{i = 1}^{n }
{y_i T\left( {\mathbf x_i ,\mathbf z } \right)}
\end{equation}
Then, the sign of $f_{\rm inter} \left( \mathbf z
\right)$ decides  the class label of $\mathbf z$.

\section{Joint Intermodal and Intramodal Label Transfers}

In addition to the above inter-modal label transfer model, we can transfer
the image labels from the training set $\mathcal T$ directly to annotate a test image $\mathbf z$.
Formally, we can define the following discriminant function for the intra-modal label transfer:
\begin{equation}\label{Eq:Eq01b}
f_{\rm intra} \left( \mathbf z \right) = \sum\limits_{j = 1}^{m }
{y_j \alpha_j K\left( \mathbf z_j ,\mathbf z  \right)}
\end{equation}
where $\boldsymbol\alpha=[\alpha_1,\cdots,\alpha_m]^\prime$ are the real-valued coefficients for intra-modal label transfer, and
$K(\cdot,\cdot):\mathcal Z \times \mathcal Z \rightarrow \mathbb R$ is a kernel function between two images satisfying Mercer's condition (e.g., Gaussian kernel) \cite{Cristianini:BOOK00}.  This label transfer has the similar form as
the discriminant function of kernelized support vector machine \cite{Cristianini:BOOK00}, with each nonzero $\alpha_j$ corresponding to a support vector.

Combining the inter-modal label transfer (\ref{Eq:Eq01}) and the intra-modal label transfer (\ref{Eq:Eq01b}), we obtain the following discriminant function to decide the label of a test image $\mathbf z$:
\begin{equation}\label{Eq:Eq01c}
f \left( \mathbf z \right) = f_{\rm inter} \left( \mathbf z \right) + f_{\rm intra} \left( \mathbf z \right)
\end{equation}
It is worth noting that usually {\bf no} surrounding text document comes with the test image $\mathbf z$. But we can always apply the transfer function to align the test image with the text documents from the source corpus $\mathcal S$.  This solves the out-of-sample problem so the text labels can be transferred to annotate any new images.

This extends the inter-modal label transfer paradigm.
We expect the intermodal and intramodal label transfers can collaboratively annotate the test images, aggregating both the label information from both texts and images.  This can mitigate the negative transfer problem \cite{Qi:CVPR11} \cite{rosenstein2005transfer} when the text documents in the source corpus cannot properly specify the visual aspect of a test image. In this case, we expect the image labels would take over to annotate the image based on its visual appearance.  In this spirit, the intramodal label transfer component plays a role of ``watchdog" to overlook and complement the intermodal label transfer.  As to be shown in the experiment, it successfully improves the intermodal label transfer model and outperforms the compared algorithms over all the  categories for a image classification task.

The learning problem of establishing joint label transfers
boils down to learn the coefficients $\boldsymbol\alpha$, along with the transfer function that
properly explains the alignment between text and image spaces.
This overall process is illustrated intuitively in  Figure
\ref{Fig:Fig01}.  Since the  key to an effective transfer learning
process is to learn the function $T$, we need to formulate an
optimization problem which maximizes the classification accuracy
obtained from this transfer process.  First, we will first set up
the optimization problem more generally without assuming any
canonical form for $T$. Later, we will set up a {\em canonical form}
for the transfer function in the form of matrices which represent
topic spaces.  The parameters of this canonical form will be
optimized in order to learn the transfer function. We propose to optimize
the following problem to jointly learn the parameters of intermodal and intramodal functions:
\begin{equation}\label{Eq:Eq02}
\begin{array}{l}
\mathop {\min }\limits_{\alpha_j, T}  \gamma \sum\limits_{j = 1}^{m}  {\ell \left( {y_j f( \mathbf z_j ) } \right)} + \lambda \sum\limits_{k=1}^l
{\delta\left( {\mathbf x_k ,\mathbf z_k } \right)}+ \Omega \left( T \right)  \\
{\rm s.t.~~} 0\leq\alpha_j\leq C,~j=1,\cdots,m
\end{array}
\end{equation}
where (1) $\ell(\cdot)$ is the loss function of the training errors on
the labeled image set $\mathcal T$; (2) $\delta(\cdot)$ is the loss function
that measures the misalignment between the co-occurrence text-image pairs, and minimizing this loss would maximize
the value of transfer function over the co-occurrence pairs; and (3) The last term $\Omega \left( T
\right)$ regularizes the  learning of the transfer function in order  to
improve the generalization performance. In the following section, we will present the detailed forms of these loss functions and the regularizer.

In addition,  $\gamma$ and $\lambda$ are positive balancing parameters, which define
the relative importance of training data and co-occurrence pairs in
the objective function; and the bound constraint follows the conventional regularization constraint
on the coefficients in support vector machines \cite{Cristianini:BOOK00}, which is expected to yield better generalization performance.

{\noindent \bf Remark on the three data sets $\mathcal S, \mathcal C, \mathcal T$:} It is worth noting that the labelled text set $\mathcal S$ is used to propagate their labels to annotate the target images. We do not need to set the text part of the co-occurrence set $\mathcal C$ to be the same as $\mathcal S$, since the modeling of co-occurrence and the label propagation are different.  The labeled image set $\mathcal T$ can also differ from $\mathcal C$. These labeled images are used to minimize the classification errors involved in the first term of objective function (\ref{Eq:Eq02}), which is different from maximization of co-occurrence consistency in the second term.

\section{Intermodal Transfer Function}


In this section, we will design the canonical form of the transfer
function $T(\mathbf x, \mathbf z)$ in terms of underlying {\em topic spaces}.  This provides a
closed form to our transfer function, which can be effectively
optimized.  Topic spaces provide a natural intermediate
representation which can semantically link the information between
the text and images.  One of the challenges to this is that
 text and images have inherently different structure to describe
their content.  For example, text is described in the form of a
vector space of sparse words, whereas images are typically defined
in the form of feature vectors that encode the visual appearances such as color,
texture and their spatial layout.  To establish their connection, one must discover
a common structure which can be used in order to link them. A text
document usually contains several topics which describe different
aspects of the underlying concepts at a higher level. For example,
in a web page depicting a \emph{bird},
some topics such as the head, body and tail may be described in its
textual part. At the same time, there is an accompanying \emph{bird}
image illustrating them. By mapping the original text and image
feature vectors into a space with several unspecified topics, they
can be semantically linked together by investigating their
co-occurrence data. By using this idea, we construct two
transformation matrices to map text and images into a common
(hypothetical) latent topic space with dimension $r$, as in the
previous work \cite{Qi:MM09}, which makes them directly comparable.
The dimensionality is essentially equal to the number of topics.  We
note that it is not necessary to know the exact semantics of latent
topics. We only attempt to model the semantic correspondence between
the unknown topics of text and images. The learning of effective
transformation matrices (or, as we will see later,  an appropriate
function of them) is the key to the success of the semantic
translation process. These matrices are defined as follows.
\begin{equation}\label{Eq:Eq03}
\mathbf W \in \mathbb R^{r\times p}:\mathcal X\subset\mathbb R^p  \to \mathbb R^r ,\mathbf x  \mapsto \mathbf W \mathbf x
\end{equation}
\begin{equation}\label{Eq:Eq04}
\mathbf V \in \mathbb R^{r\times q}:\mathcal Z\subset\mathbb R^q  \to \mathbb R^r ,\mathbf z  \mapsto \mathbf V \mathbf z
\end{equation}
The transfer function is defined as a  function of the  source and
target instances by computing the inner product in our hypothetical
topic space, with a nonlinear hyperbolic tangent activation $\tanh(\cdot)$
\begin{equation}\label{Eq:Eq05}
\begin{aligned}
T\left( \mathbf x, \mathbf z\right) &= \tanh(\left\langle \mathbf W\mathbf x ,\mathbf V\mathbf z  \right\rangle)\\
&= \tanh(\mathbf x^{\prime} \mathbf W^{\prime} \mathbf V \mathbf z)  = \tanh(\mathbf x^{\prime} \mathbf S\mathbf z)
\end{aligned}
\end{equation}
Here  $\left\langle { \cdot , \cdot } \right\rangle$ and $^\prime$
denote the inner product and transpose operations respectively.
Clearly, the choice of the transformation matrices (or rather the
product  matrix $\mathbf W^{\prime} \mathbf V$) impacts the transfer
function $T$ directly.   Therefore, we will use the notation  $\mathbf S$ in
order to  briefly denote the matrix $\mathbf W^{\prime} \mathbf V$.
Clearly, it suffices to learn this product matrix  $\mathbf S$ rather than
the two transformation matrices separately. The above definition of
the matrix $\mathbf S$ can be used to rewrite the inter-modal label transfer function as
follows:
\begin{equation}\label{Eq:Eq07}
f_{\rm inter} \left( \mathbf z \right) = \sum\limits_{i = 1}^{n }
{y_i \tanh(\mathbf x_i^\prime \mathbf S \mathbf z) }
\end{equation}

\section{Objective Problem}
Putting together with the intermodal and intramodal label transfer formula in (\ref{Eq:Eq07}) and (\ref{Eq:Eq01b}),
we define the discriminant function $f(\mathbf z)=f_{\rm inter}(\mathbf z)+f_{\rm intra}(\mathbf z)$ which can be substituted in the
objective function of the optimization problem (\ref{Eq:Eq02}) for learning the transfer
function. In addition,  we use the conventional squared norm to
regularize the transfer function $T$ on two transformations respectively:
$$\Omega \left( T \right) = \dfrac{1}{2}\left( {\left\| \mathbf W
\right\|_F^2  + \left\| \mathbf V \right\|_F^2 } \right)$$ Here,
the expression $\left\| \cdot \right\|_F$  represents the Frobenius
norm. Then,  we can use the aforementioned substitutions in order
to rewrite the objective function of Eq. (\ref{Eq:Eq02}) as follows:
\begin{equation}\label{Eq:Eq06}
\begin{aligned}
\mathop {\min }\limits_{\mathbf S=\mathbf W^{\prime} \mathbf V, \boldsymbol\alpha} & \gamma \sum\limits_{j = 1}^{m} {\ell \left( y_j f( \mathbf z_j  )  \right)} + \lambda \sum\limits_{k=1}^l {\delta\left( \mathbf x_k^{\prime} \mathbf S \mathbf z_k \right)}
 \\
&+ \dfrac{1}{2}\left( {\left\| \mathbf W \right\|_F^2  + \left\| {\mathbf V} \right\|_F^2 } \right)\\
{\rm s.t.~~} & 0\leq \alpha_j\leq C,~j=1,\cdots,m
\end{aligned}
\end{equation}

The goal is to determine the value of $\mathbf S$, which optimizes the
objective function  in  Eq. (\ref{Eq:Eq06}). We note that this
objective function  is not jointly convex in $\mathbf W$ and $\mathbf V$. This implies that the optimum
value of $\mathbf S$  may be hard to find  with the use of straightforward
gradient descent techniques, which can easily get stuck in local
minima. Fortunately, it is possible to learn $\mathbf S$ directly from Eq.
(\ref{Eq:Eq06}) by the trace norm as in \cite{Srebro:NIPS05}
\cite{Amit:ICML07}.  It is defined as follows:
\begin{equation}\label{Eq:Eq08}
\left\| \mathbf S \right\|_\Sigma   = \mathop {\inf }\limits_{\mathbf S = \mathbf W^{\prime} \mathbf V } \frac{1}{2}\left( {\left\| \mathbf W \right\|_F^2  + \left\| \mathbf V \right\|_F^2 } \right)
\end{equation}
The trace norm is a convex function of $\mathbf S$, and can be computed as
the sum of its singular values.   The  trace norm is different from
the conventional squared norm for regularization purposes,  and is
actually  a surrogate of matrix rank \cite{Cai08}, and minimizing it
can limit the dimension $r$ of the  topic space.  In other words,
minimizing the trace norm results in the fewest topics to explain
the correspondence between text and images.    This implies that
concise semantic transfer  with fewer topics is more effective than
tedious translation on cross-domain correspondence between text and
images, as long as the learned transfer function complies with the
observations (i.e., the co-occurrence and auxiliary data). This is
consistent with the parsimony principle, which states preference for
the least complex translation model.   A parsimonious choice is also
helpful in avoiding overfitting problems which may arise in
scenarios where the number of auxiliary training examples are small.

The  objective function in Eq. (\ref{Eq:Eq06}) can be rewritten as
follows with the use of the trace norm:
\begin{equation}\label{Eq:Eq09}
\mathop {\min }\limits_{\mathbf S,~0\preceq\boldsymbol\alpha\preceq C} \gamma \sum\limits_{j = 1}^{m} {\ell \left( y_j f( \mathbf z_j  )  \right)} + \lambda \sum\limits_{k=1}^l {\delta\left( \mathbf x_k^{\prime} \mathbf S \mathbf z_k \right)}  + \left\| \mathbf S \right\|_\Sigma
\end{equation}
We note that this objective function has
has a number of properties, which can be leveraged for optimization
purposes. In the next section, we discuss the methodology for
optimization of this objective function.

\subsection{Joint Optimization Algorithm}

%
In order to optimize the objective function above, we first need to
decide which functions are used for $\ell(\cdot)$ and $\delta(\cdot)$
in Eq. (\ref{Eq:Eq09}).

Recall that these functions are used  to
measure compliance with the observed co-occurrence and the margin of
discriminant functions $f(\cdot)$ on  the auxiliary data set,
respectively.  In this case, we use the hinge loss $\ell(\tau)\triangleq(1-\tau)_+$ for the loss function over the training set, where $(\cdot)_+$
denotes the positive component. We choose the hinge loss here because it has been shown to be more robust to the noisy outliers of training examples.
Clearly, minimizing the hinge loss tends to maximize the margin $y_j f(\mathbf z_j)$.

On the other hand, in compliance with the use of hyperbolic tangent activation in Eq. (\ref{Eq:Eq05}), we choose $\delta(a_k)\triangleq -\log\dfrac{1}{2}(1+\tanh(a_k))=\log(1+\exp(-2a_k))$ in the objective function (\ref{Eq:Eq09}) with $a_k$ denoting $\mathbf x_k^\prime \mathbf S \mathbf z_k$. This choice of the loss function essentially uses the logistic loss to measure the misalignment made by the transfer function between a co-occurrence pair of $\mathbf x_k$ and $\mathbf z_k$.  Minimizing this logistic loss tends to maximize the values of the transfer function over the co-occurrence pairs.

The aforementioned substitutions instantiate the objective
function (\ref{Eq:Eq09}) which is nonlinear in $\mathbf S$ and $\boldsymbol\alpha$.
One possibility for optimizing an objective function of the form
represented in Eq. (\ref{Eq:Eq09}) is to use the method of Srebro et
al. \cite{Srebro:NIPS05}. The work showed that the dual
problem can be optimized by the use of semi-definite programming
(SDP) techniques. Although many off-the-self SDP solvers use
interior point methods and return a pair of primal and dual optimal
solutions \cite{Boyd:Book04}, they do not scale  well with the size
of the problem.  The work in \cite{Amit:ICML07} proposes a gradient
based method which replaces the non-differentiable trace norm with a
smooth proxy.  But the smoothed approximation to $\|\mathbf S\|_\Sigma$ may
not guarantee that the obtained minima still correspond to fewest
topics for label transfer.

Alternatively, a proximal gradient
method is proposed in \cite{Toh:oo09} to minimize such  non-linear
objective functions with the use of a trace norm regularizer. We
will use such an approach to optimize over $\mathbf S$ and $\alpha_j$ in an alternating fashion in this paper. In order to represent the
objective function of Eq. (\ref{Eq:Eq09}) more succinctly, first we
introduce the optimization over $\mathbf S$ , and we define the function $F\left( \mathbf S \right)$ as follows.
\begin{equation}\label{Eq:Eq12}
F\left( \mathbf S \right) =  \gamma \sum\limits_{j = 1}^{m} {\ell \left( {y_j f \left( \mathbf z_j \right)} \right)} + \lambda \sum\limits_{ k=1}^l {\delta\left(\mathbf x_k^{\prime} \mathbf S \mathbf \mathbf z_k \right)}
\end{equation}

Then, the objective function of Eq. (\ref{Eq:Eq09}) can be rewritten
as $F\left(S\right)+\left\|S\right\|_\Sigma$.  In order to optimize
this objective function, the proximal gradient method quadratically
approximates it  by Taylor expansion at current value of $S=S_\tau$
with a proper coefficient $L$ as follows:
\begin{equation}\label{Eq:Eq11}
\begin{aligned}
 Q\left( {\mathbf S,\mathbf S_\tau  } \right) &= F\left( {\mathbf S_\tau  } \right) + \left\langle {\partial F\left( {\mathbf S_\tau  } \right),\mathbf S - \mathbf S_\tau  } \right\rangle   \\
&+ \dfrac{L }{2}\left\| {\mathbf S - \mathbf S_\tau  } \right\|_F^2  + \left\| \mathbf S \right\|_\Sigma   \\
&= \dfrac{L }{2}\left\| {\mathbf S - \mathbf G_\tau  } \right\|_F^2  + \left\| \mathbf S \right\|_\Sigma  + {\rm const}
\end{aligned}
\end{equation}
where $\partial F(\mathbf S)$ denotes the subgradient of $F$ at $\mathbf S$.  Here we use the subgradient because of the non-differentiability of loss function $\ell(\cdot)$ at $0$.
We can further introduce the notation $G_\tau$ in order to organize
the above expression:
\begin{equation}\label{Eq:Eq13}
\mathbf G_\tau   = \mathbf S_\tau   - L ^{ - 1} \partial F\left( {\mathbf S_\tau  } \right)
\end{equation}

The subgradient $\partial F\left( {S_\tau }
\right)$ can be computed as follows:
\begin{equation}\label{Eq:Eq14}
\begin{array}{l}
\partial F\left( {\mathbf S_\tau  } \right) =   \gamma  \sum\limits_{j = 1}^{m} y_j    \cdot \partial\ell\left( {y_j f \left( {\mathbf z_j } \right)} \right)\nabla_{\mathbf S} f(y_j f(\mathbf z_j))  \\
 + \lambda \sum\limits_{k=1}^l \left\{\partial\delta\left(\mathbf x_k^{\prime} \mathbf S_\tau \mathbf z_k\right)~ \mathbf x_k \mathbf z_k^{\prime} \right\}
 \end{array}
\end{equation}
where  $ \partial\ell$ is the subdifferential of
$\ell(\cdot)$, $\nabla_{\mathbf S} f$ is the gradient of $f$ to $\mathbf S$, and $\partial\delta$ is the derivative of logistic loss as derived before.  Then, the matrix $\mathbf S$ can be updated by minimizing $Q\left(\mathbf S, \mathbf S_\tau\right)$ with
fixed $\mathbf S_\tau$ iteratively. This  can be solved by singular value
thresholding \cite{Cai08} with a closed-form solution (see Line 4 in Algorithm \ref{Alg:Alg01}).

On the other hand, the optimization over $\boldsymbol\alpha$ can be performed by
using the gradient projection method \cite{bertsekas1999nonlinear}.  With fixed $\mathbf S$ at each iteration, each
$\alpha_j$ can be updated as
\[
\alpha_j \leftarrow \Pi_{[0,C]}(\alpha_j-\epsilon~\partial F(\alpha_j))
\]
where $\epsilon$ is a positive step size, $\Pi_{[0,C]}$ is the projection onto $[0,C]$, and
$$\partial F(\alpha_j)=\gamma\sum_{j'=1}^m{\partial\ell(y_{j'} f(\mathbf z_{j'}))K(\mathbf z_j,\mathbf z_{j'})}$$
is the subdifferential
of $F$ at $\alpha_j$.

Algorithm \ref{Alg:Alg01} summarizes the proximal gradient based
method to optimize  the expression in Eq. (\ref{Eq:Eq09}). Note that
the intermodal discriminant function $f_{\rm inter}(\mathbf z)$ is not convex as a function of $\mathbf S$,
and hence the objective function is not convex either.  But as long as the step size (i.e., $L^{-1}$) is properly set, the objective function (\ref{Eq:Eq09}) tends to decrease in each iteration, usually converging
to a stationary point (may not be a global optimum) \cite{Toh:oo09}. This is different from our previous work \cite{Qi:WWW11}, where we adopted a linear transfer function yielding a convex objective problem.  The nonlinear function has shown better performance on learning alignment between multiple modalities in literature \cite{ngiam2011multimodal}\cite{masci2014multimodal}.

{
\begin{algorithm}[tb]
   \caption{Joint Optimization for Problem (\ref{Eq:Eq09}).}\label{Alg:Alg01}
\begin{algorithmic}
   \INPUT Co-occurrence set $\mathcal C$, labeled text corpus $\mathcal S$, labeled image dataset $\mathcal T$, and balancing parameters $\lambda$ and $\gamma$.
   \STATE[1] Initialize $\mathbf S_\tau \leftarrow 0$ and $\tau\leftarrow 0$.
   \STATE[2] Initialize $\alpha_j \leftarrow 0$ for each $j$.
   \REPEAT
    \STATE[3] Set $\mathbf G_\tau   = \mathbf S_\tau   - L ^{ - 1} \partial F\left( {\mathbf S_\tau  } \right)$.
    \STATE[4] Singular Value Thresholding: $$\mathbf S_{\tau+1} \leftarrow \mathbf U\rm{diag}\left(\boldsymbol\sigma-L^{-1}\right)_+\mathbf V^\prime$$ where $\mathbf U\rm{diag}\left(\boldsymbol\sigma\right)\mathbf V^\prime$ gives the SVD of $\mathbf G_{\tau}$.
    \STATE[5] Update $\alpha_j \leftarrow \Pi_{[0,C]}(\alpha_j-\epsilon~\partial F(\alpha_j))$ for $j=1,\cdots,m$.
    \STATE[6] $\tau\leftarrow \tau+1$.
   \UNTIL Convergence or maximum iteration number achieves.
\end{algorithmic}
\end{algorithm}
}

\section{Zero-Shot Label Transfer for Unseen Classes}
The goal of zero-shot learning \cite{elhoseiny2013write,palatucci2009zero,lampert2009learning} is to build classifiers to label the unseen classes without any training image examples. However, there can be some positive examples available in text modality. In this section, we show that our cross-modal label transfer model can also be used in this setting.

Specifically, suppose that we have $n_{sc}$ {\em seen} classes with labeled training images, and our goal is to annotate the images for $n_{uc}$ unseen classes.  In addition to the images, we have the labeled text examples for both seen and unseen classes.  Then, zero-short label transfer aims to transfer the text labels  to annotate the images of unseen classes.  In principle, the same inter-modal label transfer function $f_{\rm inter}$ in Eq.~(\ref{Eq:Eq01}) is applicable in labeling the images of unseen classes, since the text labels, of no matter seen or unseen classes, can be transferred to label those images.  However, in this case, the intra-modal label transfer term $f_{\rm intra}$ will not be used any more since we cannot get access to the image labels of those unseen classes.

The learning of the inter-modal transfer function does not need to be changed to adapt to the zero-shot learning problem. However, for the sake of fair zero-shot learning scenario, we should exclude the co-occurrence text-image pairs belonging to the unseen classes from the training set. Only co-occurrence pairs of seen classes would be used to model the correlation between the text and image modalities via the inter-modal transfer function $f_{\rm inter}$.
This idea of involving pairs of seen classes has been adopted in literature \cite{kulis2011you,elhoseiny2013write} to learn the inter-modal correlations, which plays the critical role in bridging the gap across multi-modalities.

On the other hand, we note that the labeled image examples of seen classes can still be used in training the model, except that they should be treated as negative examples for the unseen classes.  These seen classes provide useful auxiliary information to exclude the regions from the feature space where the unseen classes are unlikely to be present \footnote{We assume that different classes are exclusive to each other, i.e., we consider a multi-class problem rather than a multi-label problem.  This assumption holds for many image classification problems, such as object and face recognitions.}.  This prior has been explored in \cite{elhoseiny2013write} to improve the classification accuracy for the unseen classes.

We will demonstrate the experiment result in zero-shot learning scenario in Section 8.4.

\section{Experiment}
In this section, we compare the proposed label transfer paradigm with a pure image
classification algorithm with a SVM classifier based on pure image features, along with the other existing transfer
learning methods proposed in \cite{Zhu:AAAI10}\cite{Dai:NIPS2008}\cite{Qi:WWW11}. We will show the superior results of our approach to  the other methods, with limited amount of training data.


\begin{figure}[]
    \begin{minipage}{0.5\textwidth}
      \centering
      \includegraphics[width=0.7\linewidth]{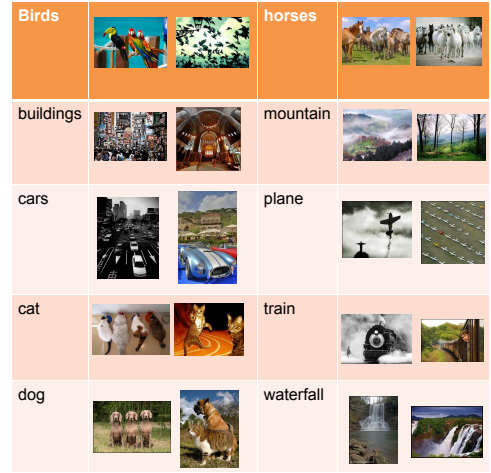}
    \end{minipage}
    \caption{Examples of images over the different categories.}\label{Fig:Fig03a} 
\end{figure}

\begin{table}[!t]
\centering \caption{The number of occurrence pairs of texts and images for each category. }\label{Tb:Tb01}
\begin{tabular}{l|c||l|c} \hline\hline
\bf Category&\bf Occurrence pairs&\bf Category&\bf Occurrence pairs\\ \hline\hline
birds&  930&    horses& 654\\
buildings&  9216&   mountain&   4153\\
cars&   728&    plane&  1356\\
cat&    229&    train&  457\\
dog&    486&    waterfall&  22006\\
\hline\hline
\end{tabular}
\end{table}

\begin{table}[!t]
\centering \caption{The number of positive and negative images for each
category.}\label{Tb:Tb01a}
\begin{tabular}{l|c|c} \hline\hline
\bf Category&\bf positive examples& \bf negative examples\\ \hline\hline
birds&  338&  349 \\
buildings&  2301&  2388\\
cars&   120&    125\\
cat&    67&    72\\
dog&    132&    142\\
horses& 263& 268\\
mountain&   927& 1065\\
plane&  509 & 549\\
train&  52 & 53\\
waterfall&  5153 & 5737\\
\hline\hline
\end{tabular}
\end{table}

\begin{table}[!t]
\centering \caption{The number of Wiki articles for each category. We collect these articles by retrieving their subcategories.}\label{Tb:Tb01c}
\begin{tabular}{l|c|l|c} \hline\hline
\bf Category&\bf Number&\bf Category&\bf Number\\ \hline\hline
birds&  82& horses & 197\\
buildings&  98 & mountain & 151\\
cars&   146 & plane &64\\
cat&    91&train &150\\
dog&    106&waterfall&249\\
\hline\hline
\end{tabular}
\end{table}

\subsection{Setting}

We  compare  the accuracy and sensitivity of our label transfer
approach with a number of algorithms below:
\begin{itemize}
\item[1.] {\em SVM \cite{Cristianini:BOOK00}.}   As the baseline, we directly train
the SVM classifiers based on the visual features extracted from images.
This method does not use any of the additional information available
in corresponding text in order to improve the effectiveness of
target domain classification. The method is also susceptible to the
case when we have a small number of test instances.
\item[2.] {\em TLRisk (Translated Learning by minimizing
Risk)\cite{Dai:NIPS2008}.} This is  another transfer learning
algorithm, which performs the translation  by minimizing risk
(TLRisk) \cite{Dai:NIPS2008}. The algorithm  transfers the text
labels to image labels via a Markovian chain. It learns a
probabilistic model to translate the text labels to image labels by
exploring the occurrence relation between text documents and images.
We note however, that such an approach does not use the topic-space
methodology which is more useful in connecting heterogeneous feature
spaces.
\item[3.]   {\em HTL (Heterogeneous Transfer
Learning)\cite{Zhu:AAAI10}:} This algorithm is the best fit to our
scenario with heterogenous spaces compared to other transfer
learning algorithms such as \cite{Raina:ICML06}\cite{Raina:ICML07}
on a homogeneous space. This method has also   been reported to
achieve superior effectiveness results. It maps
each image into a latent vector space where an implicit distance function
is formulated.  In order to do so, it also makes use of the
occurrence information between images and text documents as well as images and visual words.
 To facilitate this method into our scenario,
user tags in {\em Flickr} are extracted to construct the relational
matrix between images and tags as well as that between tags and
documents. Images are represented in a new feature space on which
the images can be classified by applying the $k$-nearest neighbor
classifier (here $k$ is set to be $3$) based on the distances in the
new space. We refer to this method as {\bf HTL}.
\item[4.] {\em Translator from Text to
Images (TTI)\cite{Qi:WWW11}:} This is our previous label transfer algorithm which only uses intermodal label transfer without considering the intramodal label transfer.  This model fails to outperform the other compared algorithms on some categories \cite{Qi:WWW11}. As aforementioned, this might be caused by the misalignment between text documents and test images.
\item[5.] {\em Joint Intermodal and Intramodal Label Transfer (I2LT):} this is the proposed approach in this paper.
\end{itemize}

In  the experiments,  a small number of
training images are randomly selected from each category as labeled instances in
$\mathcal T$ for the
classifiers. The  remaining images in each category are used for testing the performance of the  classification task. Only
a small number of training examples are used, making the problem very
challenging from the training perspective.
This process is repeated
five times. The error rate and the standard deviation for
each category is reported in order to evaluate
the effectiveness of the compared classifiers. We also use varying number of co-occurred text-image pairs
 to construct the classifier, and compare the corresponding results
with related algorithms.

In the experiments, the parameters $\lambda$, $\gamma$ (used to
decide the importance of auxiliary data and co-occurrence data from
the objective function in (\ref{Eq:Eq09})) and $C$ (used to regularize the intramodal label transfer) are selected from $\{0,
0.5, 1.0, 2.0\}$, $\{0.1, 0.5, 1.0, 2.0\}$ and $\{1.0, 2.0, 5.0, 10.0\}$, respectively. All the parameters are tuned based on a
twofold cross-validation procedure on the selected training set,
and the parameters with the best performance are selected to train
the models.

\subsection{Result on Flickr-Wiki Dataset}
\begin{table*}[!t]
\centering \caption{Comparison of error rate of different algorithms with (a) two training images (b) ten training images.  The smallest error rate for each category is in bold.   }
\subtable[Two training images]{
\begin{tabular}{l|l|l|l|l|l} \hline\hline
\bf Category&\bf SVM&\bf HTL&\bf TLRisk&\bf TTI&\bf I2LT\\ \hline\hline
birds&  0.3293$\pm$0.0105& 0.3293$\pm$0.0124& 0.2817$\pm$0.0097& {0.2738$\pm$0.0080}&\bf 0.2523$\pm$0.0042\\
buildings&  0.3272$\pm$0.0061& 0.3295$\pm$0.0041& 0.2758$\pm$0.0023& {0.2329$\pm0.0032$}&\bf 0.1985$\pm$0.0023\\
cars&   0.2529$\pm$0.0059& 0.2759$\pm0.0048$& 0.2639$\pm0.0032$& {0.1647$\pm0.0058$}&\bf 0.1326$\pm$0.0082\\
cat&    0.3333$\pm$0.0071& 0.3333$\pm$0.0060& {0.2480$\pm$0.0109}& 0.2525$\pm$0.0083&\bf 0.2256$\pm$0.0024\\
dog&    0.3694$\pm$0.0031& 0.3694$\pm$0.0087& 0.2793$\pm$0.0161& {0.252$\pm$0.0092}&\bf 0.2415$\pm$0.0074\\
horses& 0.25$\pm$0.0087&   0.3$\pm$0.0050&    0.2679$\pm$0.0069& {0.2$\pm$0.0015}&\bf 0.1879$\pm$0.0093\\
mountain&   0.3311$\pm$0.0016& 0.3322$\pm$0.0009& 0.2817$\pm$0.0021& {0.2699$\pm$0.0004}&\bf 0.2482$\pm$0.0010\\
plane&  0.2667$\pm$0.0019& {0.225$\pm$0.0006}& 0.2758$\pm$0.0006& 0.2517$\pm$0.0011&\bf 0.2044$\pm$0.0004\\
train&  0.3333$\pm$0.0084& 0.3333$\pm$0.0068& 0.2738$\pm$0.0105& {0.2099$\pm$0.0060}&\bf 0.1910$\pm$0.0014\\
waterfall&  0.2693$\pm$0.0009& 0.2694$\pm$0.0016& 0.2659$\pm$0.0020& {0.257$\pm$0.0007}&{\bf 0.2241$\pm$0.0010}\\
\hline\hline
\end{tabular}\label{Tb:Tb02}
}
\subtable[Ten training images]{
\begin{tabular}{l|l|l|l|l|l} \hline\hline
\bf Category&\bf SVM&\bf HTL&\bf TLRisk&\bf TTI&\bf I2LT\\ \hline\hline
birds&  0.2639$\pm$0.0012& 0.2619$\pm$0.0015& 0.2546$\pm$0.0018& {0.252$\pm$0.0008}&\bf 0.2314$\pm$0.0012\\
buildings&  0.2856$\pm$0.0002& 0.2707$\pm$0.0021& 0.2555$\pm$0.0014& {0.2303$\pm$0.0017}&\bf 0.2214$\pm$0.0014\\
cars&    0.3027$\pm$0.0073&0.3065$\pm$0.0030& 0.2543$\pm$0.0029& {0.2299$\pm$0.0031}&\bf 0.2025$\pm$0.0022\\
cat&    0.2755$\pm$0.0043& {0.2525$\pm$0.0038}& 0.2553$\pm$0.0028& {0.2424$\pm$0.0026}&\bf 0.2289$\pm$0.0052\\
dog&    0.2252$\pm$0.0039& 0.2343$\pm$0.0037& 0.2545$\pm$0.0031& {0.2162$\pm$0.0027}&\bf 0.2015$\pm$0.0024\\
horses& 0.2667$\pm$0.0019& {0.2500$\pm$0.0021}&  0.2551$\pm$0.0016& {0.2383$\pm$0.0013}&\bf 0.2192$\pm$0.0018\\
mountain&   0.3176$\pm$0.0010& 0.3097$\pm$0.0003& {0.2541$\pm$0.0011}&    0.2626$\pm$0.0007&\bf 0.2312$\pm$0.0004\\
plane&  0.2667$\pm$0.0009& {0.2133$\pm$0.0008}&    0.2546$\pm$0.0005& 0.2567$\pm$0.0012&\bf 0.1815$\pm$0.0004\\
train&  0.2624$\pm$0.0029& 0.2716$\pm$0.0118& 0.2552$\pm$0.0025& {0.2346$\pm$0.0031}&\bf 0.2123$\pm$0.0026\\
waterfall&  0.2611$\pm$0.0008& {0.2435$\pm$0.0009}&    0.2555$\pm$0.0016& 0.2546$\pm$0.0007&\bf 0.2252$\pm$0.0008\\
\hline\hline
\end{tabular}\label{Tb:Tb03}
}
\end{table*}

\begin{table}[!t]
\small \centering \caption{The number of topics (i.e., the rank of matrix $S$) used for learning the transfer function in topic space from $2,000$ co-occurrence pairs with two and ten training examples. }\label{Tb:Tb03a}
\vspace{2mm}
\begin{tabular}{l|c|c} \hline\hline
\bf Category&\bf Two examples& \bf Ten examples\\ \hline\hline
birds&  15& 28\\
buildings&  75& 96\\
cars&   7& 16\\
cat&    11& 15\\
dog&    7&  14\\
horses& 5&  7\\
mountain&   5&  9\\
plane&  17& 21\\
train&  8&  6\\
waterfall&  19& 28\\
\hline\hline
\end{tabular}
\end{table}
\begin{figure}[!t]
\begin{center}
\includegraphics[width=0.7\linewidth]{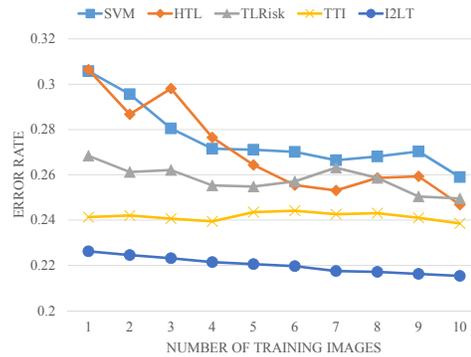}
\end{center}
 \caption{Average error rate of different algorithms with varying number of training images.}\label{Fig:Fig03}
\end{figure}

\begin{figure}[!t]
\begin{center}
\includegraphics[width=0.7\linewidth]{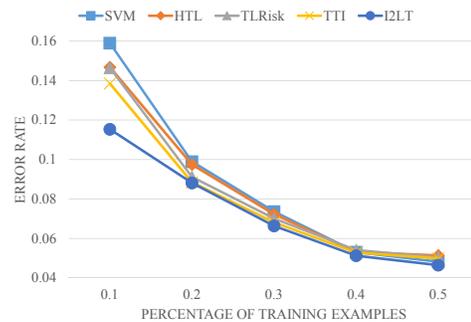}
\end{center}
 \caption{Average error rate of different algorithms with various percentage of training images from each concept, from $10\%$ to $50\%$ with an increment of $10\%$.}\label{Fig:Fig03b}
\end{figure}


The first data set is Flickr-Wiki dataset, consisting of a collection of {\em Flickr} and {\em
Wikipedia} web pages which contains rich media content
with images and their text descriptions.  We use  ten
categories to evaluate the effectiveness on the image classification
task.  To collect text and image collections for experiments, the
names of these $10$ categories are used as query keywords to retrieve
the relevant web pages from {\em Flickr} and {\em Wikipedia}. Both web sites
return many web pages in response to the submitted queries. Figure \ref{Fig:Fig03a} illustrates some
examples of retrieved images, Table
\ref{Tb:Tb01} shows the number of occurrence pairs crawled from Flickr
by using different query words, and Table \ref{Tb:Tb01c} shows the number of Wiki articles
retrieved from the subcategories of each topmost category. For example, these subcategories contain the breeds of animals (e.g., bird, horse, dog, and cat), and the list of buildings, mountains and waterfalls.

{\em Flickr} is an image sharing web site,
storing many user-shared images and their textual descriptions in the forms
of textual tags and comments.  For {\em
Wikipedia}, we have also retrieved the relevant web pages in the subcategories.
In each crawled web page, the images and the surrounding
text documents are used to learn the alignment between text and
images.  It is worth noting that these co-occurrence pairs used to align the image and text modalities
do {\em not} contain any labeled images in the training set. In other words, no images in the co-occurrence pairs are labeled, and hence, these pairs are unlabeled. In fact, in our algorithm, we do not need the labels of these pairs to learn label transfer.  These unlabeled pairs are only used to model the correlation between the two modalities.

For images, visual features are extracted to describe these images.
For the sake of fair comparison,
we use the same vocabulary of visual words to
represent images as those used by the compared algorithms in previous work \cite{Qi:WWW11}. These  include the  $500$ dimensional bag of visual-words (BOVW)
based on SIFT descriptors \cite{Lowe:IJCV04}. For the text
documents, we normalize the textual words by removal of stop words and stemming, and use their frequencies as textual features.
For each category, the images are manually annotated to collect the
ground truth labels for training and evaluation as shown in Table
\ref{Tb:Tb01a}.  Nearly the same number of background images are
collected  as the negative examples.  These background images do not
contain the objects of the categories.  It is worth noting that these image categories are
not exclusive which means that one image can be annotated by more than one category.

First, in Figure \ref{Fig:Fig03} and \ref{Fig:Fig03b}, we report the performances of different algorithms
with varying
numbers of training images.  For each category,  the same number of
images from the background images are used as the negative examples.
Then average error rate is shown to
evaluate the performance for image classification tasks. To learn
the transfer function, $2,000$ co-occurrence pairs are collected to
 learn the alignment between texts and images for label transfer.
 Since each image can be assigned
more than one label, the error rate is computed in binary-wise fashion.

We note that as shown in Figure \ref{Fig:Fig03}, a small number of
 training images is the most interesting case for
our algorithm, because it handles the challenging cases when an image category do not have much past
 labeling information for the
classification process. In order to validate this point, in Figure \ref{Fig:Fig03}, we  compare the average error
rates over all categories with varying number of auxiliary training
examples.
 It demonstrates the advantages of our methods when there
are an  extremely small number of training images.
This confirms our earlier assertion that our
approach can work even in the paucity of auxiliary training
examples, by exploring the correspondence between text and images.
In Tables \ref{Tb:Tb02} and \ref{Tb:Tb03}, we compare the error rate of
different algorithms for each category with two and ten auxiliary
training images respectively.     We note that Table \ref{Tb:Tb02} (a)
shows the results with a {\em extremely smaller number of
training images, and the proposed scheme outperforms
the compared algorithms on all the categories.}
If we continue to increase the number of training examples to a large enough level,
as shown in Figure \ref{Fig:Fig03b}, the advantage achieved by the label transfer algorithm
gradually diminishes.  This is expected since with sufficiently training examples, there is no need to
leverage the cross-modal labels to enhance the classification accuracy.

Also, Table \ref{Tb:Tb03a} lists the
number of topics (i.e., the rank of matrix $\mathbf S$) used for learning the transfer function
in topic space from $2000$ co-occurrence pairs with two and ten training examples.  It shows that for most of categories with only a
small number of topics, the learned label transfer model works very
well.  This also provides evidence of the advantages of the parsimony
principle in semantic translation.  However, this criterion is not absolute or unconditioned, but with the premise that the observed training examples and co-occurrence pairs can be fit by the learned model.  For complex categories with many aspects, it often uses more topics to establish the correspondence between the heterogeneous domains. For example, as the appearances of ``buildings" vary largely with lots of variants, more topics are needed to explain the correspondence between these variants than the categories with relatively uniform appearances. But as long as the training data can be explained, the models with fewer topics are preferred for the improved generalization performance.

\subsection{Result on NUS-WIDE Dataset}

The second dataset we use to evaluate the algorithm is NUS-WIDE \cite{nus-wide-civr09}, which is a real-world
image dataset that contains $269,648$ images downloaded
from Flicker. Each image has a number of textual
tags and is labeled with one or more image concepts
out of $81$ concepts.
The $186, 577$ image-text pairs belonging to the $10$
largest concepts are selected as co-occurrence pairs.
Similar to the above Flickr-Wiki dataset, the images are represented
by $4096$-D Convolutional Neural Network (CNN) features by AlexNet \cite{alexnet} and the
image tags are represented by $1000$-D word occurrence
feature vectors.


\begin{figure*}[!t]
      \centering
      \includegraphics[width=0.6\linewidth]{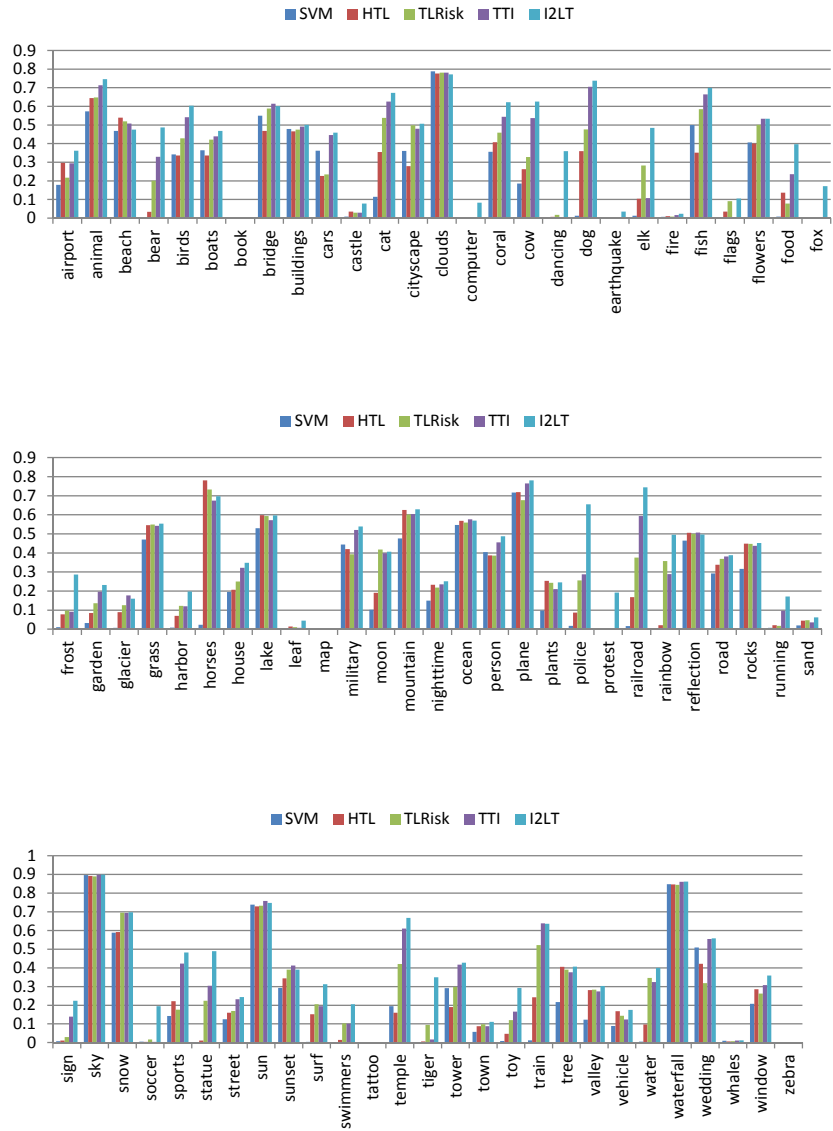}
    \caption{Comparison of Average Precision (AP) for $81$ concepts on NUS-WIDE dataset. The plot is better viewed in color.}\label{Fig:Fig06} 
\end{figure*}

\begin{table}[!t]
\centering \caption{Comparison of Mean Average Precision (MAPs) on NUS-WIDE dataset.}\label{Tb:Tb05}
\vspace{2mm}
\begin{tabular}{l|c} \hline\hline
\bf Algorithms&MAPs\\ \hline\hline
SVM&0.2078\\
HTL&0.2563\\
TLRisk&0.3048\\
TTI&0.3418\\
I2LT&0.4042\\
\hline\hline
\end{tabular}
\end{table}

Figure~\ref{Fig:Fig06} plots the comparison of Average Precision (AP) for $81$ concepts on NUS-WIDE dataset.  Average Precision measures how well an algorithm ranks the positive examples higher than the negative examples \cite{zhutech2014precision}.  It is a widely-used metric in comparing between different classification algorithms especially with an imbalanced sets of the positive and negative examples.  From the result, we can find that on 67 out of 81 concepts, the proposed I2TL outperforms the other compared algorithms. Table~\ref{Tb:Tb05} compares the Mean Average Precision (MAPs) over 81 concepts on NUS-WIDE dataset.

\begin{table}[!t]
\caption{Comparison of CCA-type cross-modal retrieval models and the proposed I2LT algorithm.
The performance reported in the table is Precision@20. The results are an average from five random splits of database/query splits, and the standard deviations are on an interval of $[0.52\%-1.34\%]$. }\label{Tb:Tb06}
\centering
\vspace{2mm}
\begin{tabular}{l|c|c} \hline\hline
\bf Algorithms&I2I&T2I\\ \hline\hline
CCA(V+T)&42.44&42.37\\
CCA(V+T+K)&48.06&50.46\\
CCA(V+T+C)&44.03&43.11\\
I2LT+BOVW&56.34&56.85\\
I2LT+CNN & 68.72&69.20\\
\hline\hline
\end{tabular}
\end{table}

It is worth noting that the learned intermodal label transfer function measures the cross-modal relevance.  It can be used to retrieve the relevant the images given a query of text description, and vice versa.  Thus, we test the cross-modal retrieval with the learned transfer function. Specifically, following the experimental setup in \cite{gongijcv2013multiview}, we consider two scenarios. (1) The Image to Image (I2I) search, i.e., an image is used as a query to search the relevant images with the same label; (2)
The Text to Image (T2I) search, i.e., the input query is a text description and the output is a list of relevant images.  We can also perform an Image to Text (I2T) search, where an image is used as input query to search for the relevant text descriptions. However, the I2T result on NUS-WIDE was not reported in \cite{gongijcv2013multiview}. For the sake of a straight comparison, we skip the I2T search in this paper too.

We follow the same evaluation protocol as \cite{gongijcv2013multiview}: from the test set, $1,000$ samples are randomly sampled and used as the queries,
 $1,000$ as the validation set, and the remaining ones are retrieved. A retrieved output is considered as relevant to an input query if they have the same label.  The experiments are repeated five times, and we report the top-20 precision averaged over the five random database/query splits.

Table~\ref{Tb:Tb06} compares the retrieval performances by I2LT and the other three CCA variants. Among them,  CCA(V+T) refers to the two-view baseline model based on both visual and text features; CCA(V+T+K) refers to the three-view CCA model with visual, text and supervised semantic information; and CCA(V+T+C) refers to the three-view model with unsupervised third view on automatically generated word clusters.  More details about these three models can be found in \cite{gongijcv2013multiview}.  In testing I2I search,
image features are projected into the CCA space and the learned I2LT space ({\em cf.} Eq.~(\ref{Eq:Eq04})) respectively, and then we use them to retrieve the most relevant images from the dataset.  For the fair comparison with these CCA variants trained with the original BOVW features, we report the retrieval accuracies by I2LT with BOWV features and CNN features in the table.

\subsection{Zero-Shot Label Transfer on CUB200 and Oxford Flower-102 Datasets}
We used two datasets to test the algorithm for the zero-shot label transfer. The first one is CUB200 Birds dataset \cite{WelinderEtal2010} which consists of 200 species of birds in $6033$ images.  The corresponding wikipedia articles are collected by using the name of these birds as query keywords, ending up with $200$ articles as the text descriptions \cite{elhoseiny2013write}. The second dataset is Flower102 with 102 classes of flowers in $8189$ images \cite{nilsback2008automated}. Different from CUB200, the text articles, one for each flower class, are collected not only from Wikipedia, but also from Plant Database, Plant Encyclopedia, as well as BBC articles \cite{elhoseiny2013write}.

Both datasets extracted $2569$ dimensional Classme features as an intermediate semantic representation of the input images.  For the text modality, TF-IDF (Term-Frequency and Inverse Document Frequency) features are extracted from each article, followed by reducing $8875$-dimensional TF-IDF features to $102$ dimension with Cluster Latent Semantic Indexing (CLSI) algorithm.
The resultant dataset with text descriptions is publicly available \cite{elhoseiny2013write} \footnote{\url{https://sites.google.com/site/mhelhoseiny/computer-vision-projects/Write_a_Classifier}}.

Five-fold cross validation over the was adopted to test the algorithm, where 4/5 classes were used as seen classes and the other 1/5 of classes as unseen ones. Then the datasets are split into training and test sets according to the seen and unseen classes, where the images and the corresponding articles of thoese seen classes constitute the co-occurrence pairs.  The five-fold cross-validation over the seen classes is used to decide the hyper-parameters.  Following \cite{elhoseiny2013write}, we report the average AUC (Area Under ROC Curve) over five-fold cross-validation to evaluate the performance.

We considered four state-of-the-art zero-shot learning algorithms as baselines, namely (1) Gaussian Process Regressor (GPR) \cite{rasmussen2006gaussian}, (2) Twin Gaussian Process (TGP) \cite{bo2010twin}, (3) Nonlinear Asymmetric Domain Adaptation (DA) \cite{kulis2011you}, as well as (4) WAC (Write A Classifier) \cite{elhoseiny2013write}.

We report the comparative results on the two datasets in Table~\ref{Tb:Tb08}.
We can see that the proposed algorithm outperforms the others in terms of average AUC. The performance improvement is partly attributed to the fact that the proposed approach prefers the concise label transfer model by imposing the trace norm regularizer.  This preference plays an important role considering that only very rare positive examples are available for unseen classes in text and image modalities (There exists no image examples for zero-shot learning!).  With extremely rare examples, adopting a concise cross-modal transfer model can minimize the over fitting risk effectively as both modalities have much high dimensionality of feature representations.  Actually, the resultant label transfer matrices $\mathbf S$ are only of rank $35 \pm 3$ on CUB200 dataset and of rank $28 \pm 5$ on Flower102 dataset over five-fold cross-validation.

We also compare the classification accuracies with various types of output embedding models \cite{akata2015evaluation} on the extended CUB dataset in Table~\ref{Tb:Tb09}.  This dataset extends CUB200 dataset to have $11,788$ images from $200$ bird species.
For a fair comparison, the same zero-shot split as in \cite{akata2013label}\cite{akata2015evaluation} is used, where 150 classes are used for the
training and validation, and the remaining $50$ disjoint classes are used for testing.
The average
per-class accuracy is reported on the test set for each compared algorithm. From the comparison, we can find the proposed algorithm outperforms the compared types of embedding algorithms.  This can be attributed to the proposed algorithm which does not only use input and output embeddings to learn the transfer function, but also applies the learned transfer function to combine multiple labels of source texts to annotate the target images.  On the contrary, these existing embedding models only output the compatibility between an input-output pair, without exploring the joint use of multiple source labels to predict on a target image.

Here, we wish to make an additional note on the training of the proposed label transfer model in the zero-shot scenario. In the experiment, we enforce that the label transfer matrix $\mathbf S$ is shared across seen and unseen classes. This is possible because the transfer matrix is class-independent, since it aims to capture the inter-modal correlation between images and their corresponding articles no matter which classes they belong to.  In this sense, in the training phase, each seen class of images in the training set can also be labeled by transferring the corresponding text labels with the shared transfer matrix learned by minimizing such label transfer errors as in Eq. (\ref{Eq:Eq02}). In this way, we fully explore the image labels of seen classes in the training set to learn the shared transfer matrix. This does not violate the zero-shot assumption that no image labels of unseen classes should be involved in the training algorithm.  The experiment results also show that, without this training strategy,
 the proposed approach only achieved an average AUC of $0.64$ and $0.70$ on CUB200 and Flower102 by learning a separate transfer matrix for each unseen class, justifying this transfer matrix sharing strategy.

\begin{table}[!t]
\caption{Comparison of Zero-Shot classifiers on CUB200 and Flower102. Average AUC and its standard deviation are reported.}\label{Tb:Tb08}
\centering
\vspace{2mm}
\begin{tabular}{l|c|c} \hline\hline
\bf Algorithms&CUB200&Flowers102\\ \hline\hline
GPR&0.52$\pm$0.001&0.54$\pm$0.02\\
TGP&0.61$\pm$0.02&0.58$\pm$0.02\\
DA&0.59$\pm$0.01&0.62$\pm$0.03\\
WAC&0.62$\pm$0.02&0.68$\pm$0.01\\
Our approach&{\bf 0.67$\pm$0.02}&{\bf 0.73$\pm$0.01}\\
\hline\hline
\end{tabular}
\end{table}

\begin{table}[!t]
\caption{Comparison of Zero-Shot classifiers on the extended CUB dataset. We compare with the Structured Joint Embedding (SJE) on various types of output embeddings. Classification accuracies are reported.}\label{Tb:Tb09}
\centering
\vspace{2mm}
\begin{tabular}{l|c} \hline\hline
\bf Algorithms&Extended CUB\\ \hline\hline
Word2Vec($\varphi^{\mathcal W}$)&28.4\%\\
GloVe($\varphi^{\mathcal G}$)&24.2\%\\
Bag-of-Words($\varphi^{\mathcal B}$)&22.1\%\\
WordNet($\varphi^{\mathcal H}$)&20.6\%\\
human($\varphi^{0,1}$)&37.8\%\\
human($\varphi^{\mathcal A}$)&50.1\%\\
Our approach&{\bf 55.3\%}\\
\hline\hline
\end{tabular}
\end{table}

\subsection{Impact of the Size of Co-occurrence Pairs}
\begin{figure}[!t]
    \begin{minipage}{0.5\textwidth}
      \centering
      \includegraphics[width=0.65\linewidth]{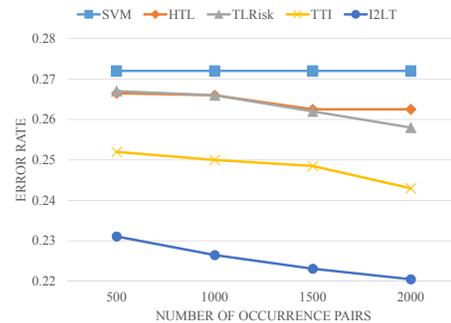}
    \end{minipage}
    \caption{Average error rate of different algorithms with varying number of image-text pairs to learn the intermodal transfer function.}\label{Fig:Fig05} 
\end{figure}

These above results are obtained by using $2,000$ pairs of co-occurred text and images.
We know the number of co-occurrence text-image pairs play an
important role to align the heterogeneous modalities. Therefore, it is instructive to
examine the effect of increasing the pair numbers.  In
Figure \ref{Fig:Fig05}, we compare the error rates of different
algorithms with varying numbers of text-image pairs. The number of
pairs is illustrated on the horizontal axis, whereas the error rate
is illustrated on the vertical axis.  As we can see,  the error rate
of the proposed I2LT algorithm decreases with an increasing
number of pairs because more information is exploited to align text and image domains. We also note that its improvement is more
significant than other algorithms when more text-image pairs are
involved.  This shows that I2LT is more resistant against
the noisy co-occurrence pairs of texts and images by
 jointly modeling the relevance of training labels between the texts and the images
 used for label transfer.  It also demonstrates the advantage of I2TL over the other algorithms.

\subsection{Computational Cost}

\begin{table}[!t]
\small \centering \caption{The training and testing time with $2000$ co-occurrence pairs and $10$ training examples (in seconds).}\label{Tb:Tb04}
\vspace{2mm}
\begin{tabular}{l|c|c} \hline\hline
\bf Algorithms&\bf Training &\bf Testing\\ \hline\hline
SVM&0.6&8.2\\
HTL&22.6&24.1\\
TLRisk&17.2&14.5\\
TTI&25.3&26.1\\
I2LT&25.4&29.3\\
\hline\hline
\end{tabular}
\end{table}

Finally we compare the computational costs made by different algorithms.  All the algorithms are conducted on the same cluster server, equipped with Intel Xeon 2.5 GHz 12-Core CPU, and $128$ GB physical memory. Table \ref{Tb:Tb04} shows the computing time to train and test with the different models.  It is shown that SVM is the fastest model to train since it does not involve any labeled text corpus.  TLRisk is the second fastest model to train, and the other three models are trained in the comparable time since all of them spend most of time on constructing intermediate representation to transfer the labels. For test, I2LT uses the longest time because it has to transfer the labels from the intermodality as well as intramodality.  But the longer time is compensated by the more accurate test results as shown above.

\section{Conclusion}
In this paper, we presented a method  to jointly transfer labels within and across
modalities for an effective image
classification model. This method is designed in order to alleviate the
dual issues of scare labels and high semantic gaps which are
inherent for the images. The label transfer process is designed with
the development of  a transfer function, which can convert the  labels
from text to images  effectively.
We show that the transfer function
can be learned from the co-occurrence pairs of texts and images as well as
a small size of training images. We follow the parsimonious principle to develop a common
representation to align texts and images with as few topics as possible in the label transfer process.
For prediction, we {\bf do not} assume that a test image comes with any text description,
and the labels of  the text corpus can be
propagated to annotate the test image by the learned transfer function.
We show superior
results of the proposed algorithm for the image classification task
as compared with state-of-the-art heterogeneous transfer learning
algorithms.

\section*{Acknowledgement}
The first author was partly supported by NSF grant 16406218.  We also would like to thank the anonymous reviewers for bringing the zero-shot learning problem into our attention, which inspires us to study the applicability of the proposed approach to this problem.
%
%

%

%

{
\bibliographystyle{abbrv}
\bibliography{english,sigproc}
}


\begin{IEEEbiography}[{\includegraphics[width=1in,height=1.25in,clip,keepaspectratio]{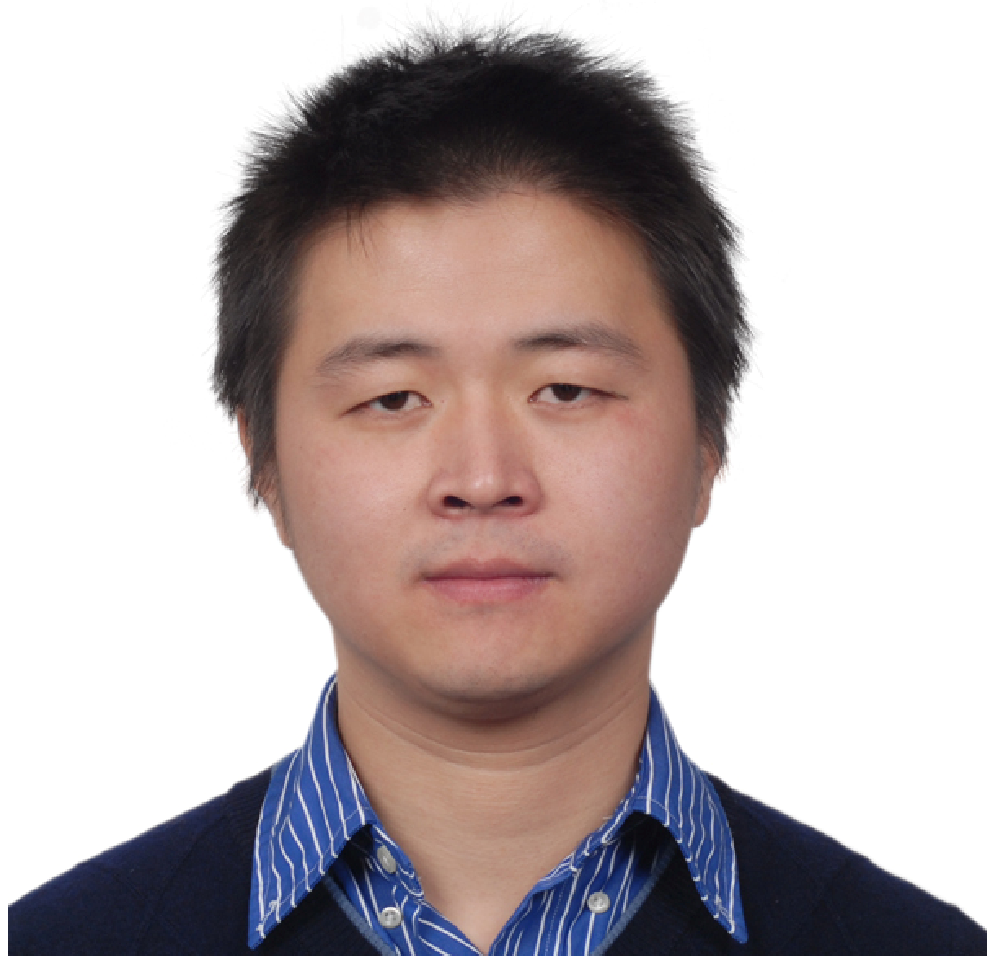}}]{Guo-Jun Qi}
Dr. Guo-Jun Qi is an assistant professor in the Department of Computer Science at University of Central Florida.  He
received the Ph.D. degree in Electrical and Computer Engineering from the University of Illinois at Urbana-Champaign.
His research interests include pattern recognition, machine learning,
computer vision and multimedia.  He was the co-recipient of the best student paper award
in IEEE Conference on Data Mining (2014), and the recipient of the best
paper award (2007) and the best paper runner-up (2015) in the ACM International Conference on Multimedia.  He has served or will serve as program co-chair of MMM 2016, an area chair of ACM Multimedia (2015, 2016),
a senior program committee member of ACM CIKM 2015 and ACM SIGKDD 2016, and program committee members or reviewers for the conferences and journals in the fields of computer vision, pattern recognition,
machine learning, and data mining. Dr. Qi has published over 60 academic papers in these areas.
He also (co-)edited the two special issues on IEEE transactions on multimedia and IEEE transactions on big data.
\end{IEEEbiography}

\begin{IEEEbiography}[{\includegraphics[width=1in,height=1.25in,clip,keepaspectratio]{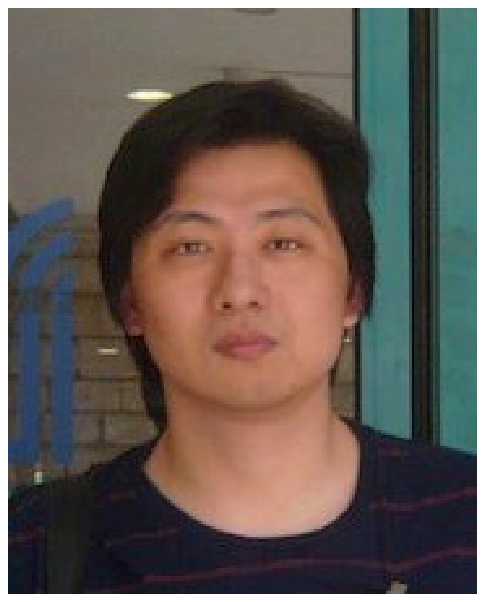}}]{ Wei Liu}
Dr. Wei Liu received the M.Phil. and Ph.D. degrees in electrical engineering from Columbia University, New York, NY, USA in 2012.
Currently, he is a research staff member of IBM T. J. Watson Research Center, Yorktown Heights, NY, USA, and holds adjunct faculty positions
at Rensselaer Polytechnic Institute and Stevens Institute of Technology. He has been the Josef Raviv Memorial Postdoctoral Fellow at
IBM T. J. Watson Research Center for one year since 2012. His research interests include machine learning, data mining, computer vision,
pattern recognition, image processing, and information retrieval. Dr. Liu is the recipient of the 2011-2012 Facebook Fellowship
and the 2013 Jury Award for best thesis of Department of Electrical Engineering, Columbia University. Dr. Liu has published over 70 papers
in peer-reviewed journals and conferences including Proceedings of IEEE, IEEE Transactions on Image Processing, NIPS, ICML, KDD, CVPR,
ICCV, ECCV, MICCAI, IJCAI, AAAI, SIGIR, SIGCHI, DCC, etc. His recent papers win CVPR Young Researcher Support Award and Best Paper Travel Award for ISBI 2014.
\end{IEEEbiography}

\begin{IEEEbiography}[{\includegraphics[width=1in,height=1.25in,clip,keepaspectratio]{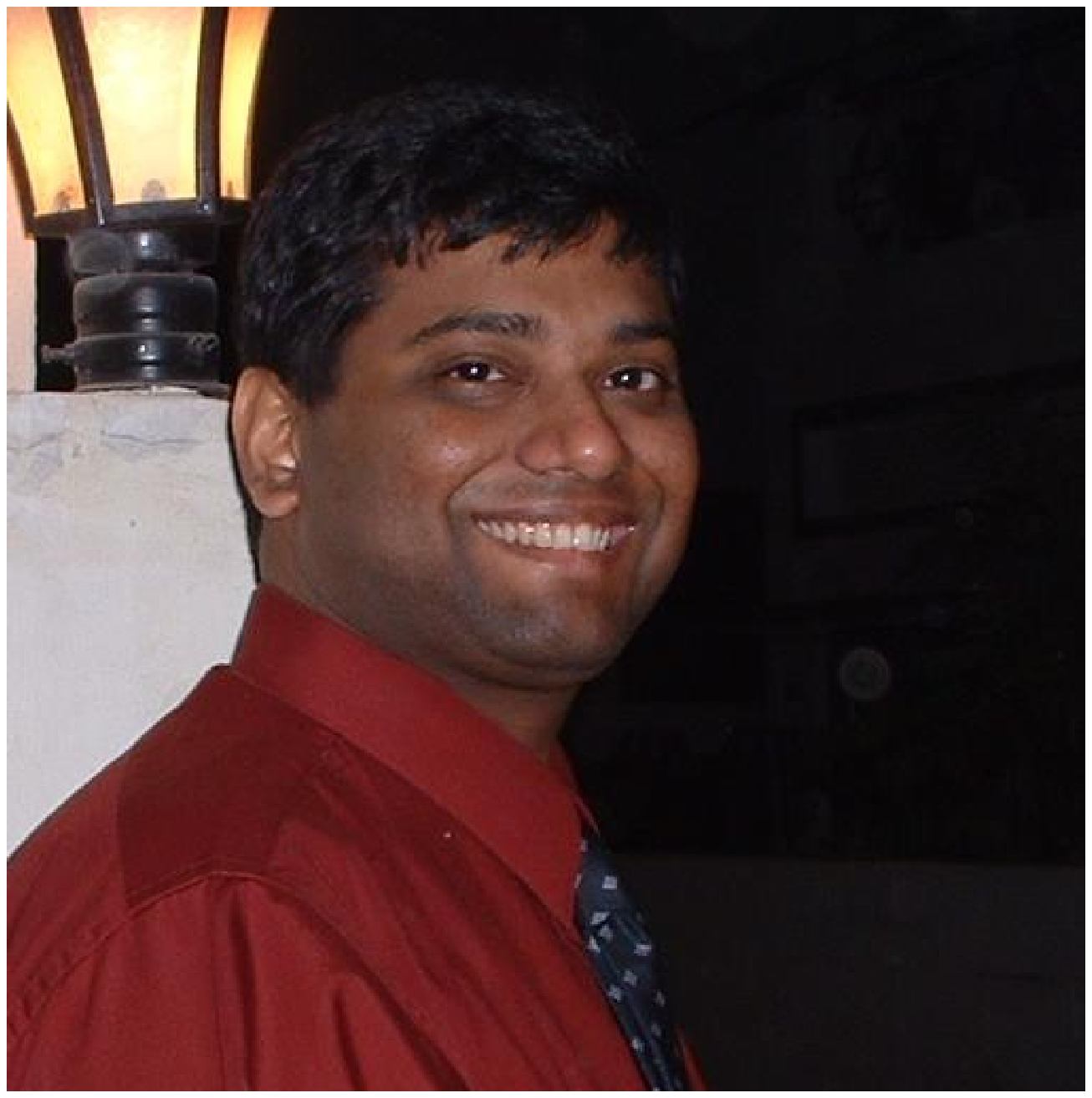}}]{Charu Aggarwal}
Dr. Charu Aggarwal is a Distinguished Research Staff Member at the IBM T. J. Watson Research Center in Yorktown Heights, New York. He completed his B.S. from IIT Kanpur in 1993 and his Ph.D. from Massachusetts Institute of Technology in 1996. He has since worked in the field of performance analysis, databases, and data mining. He has published over 135 papers in refereed conferences and journals, and has been granted over 50 patents. He has served on the program committees of most major database/data mining conferences, and served as program vice-chairs of the  SIAM Conference on Data Mining , 2007, the IEEE ICDM Conference, 2007, the WWW Conference 2009, and the IEEE ICDM Conference, 2009. He served as an associate editor of the  IEEE Transactions on Knowledge and Data Engineering Journal  from 2004 to 2008. He is an action editor of the  Data Mining and Knowledge Discovery Journal, an associate editor of the ACM SIGKDD Explorations, and an associate editor of the Knowledge and Information Systems Journal. He is a fellow of the ACM (2013) and the IEEE (2010) for "contributions to knowledge discovery and data mining techniques".
\end{IEEEbiography}

\begin{IEEEbiography}[{\includegraphics[width=1in,height=1.25in,clip,keepaspectratio]{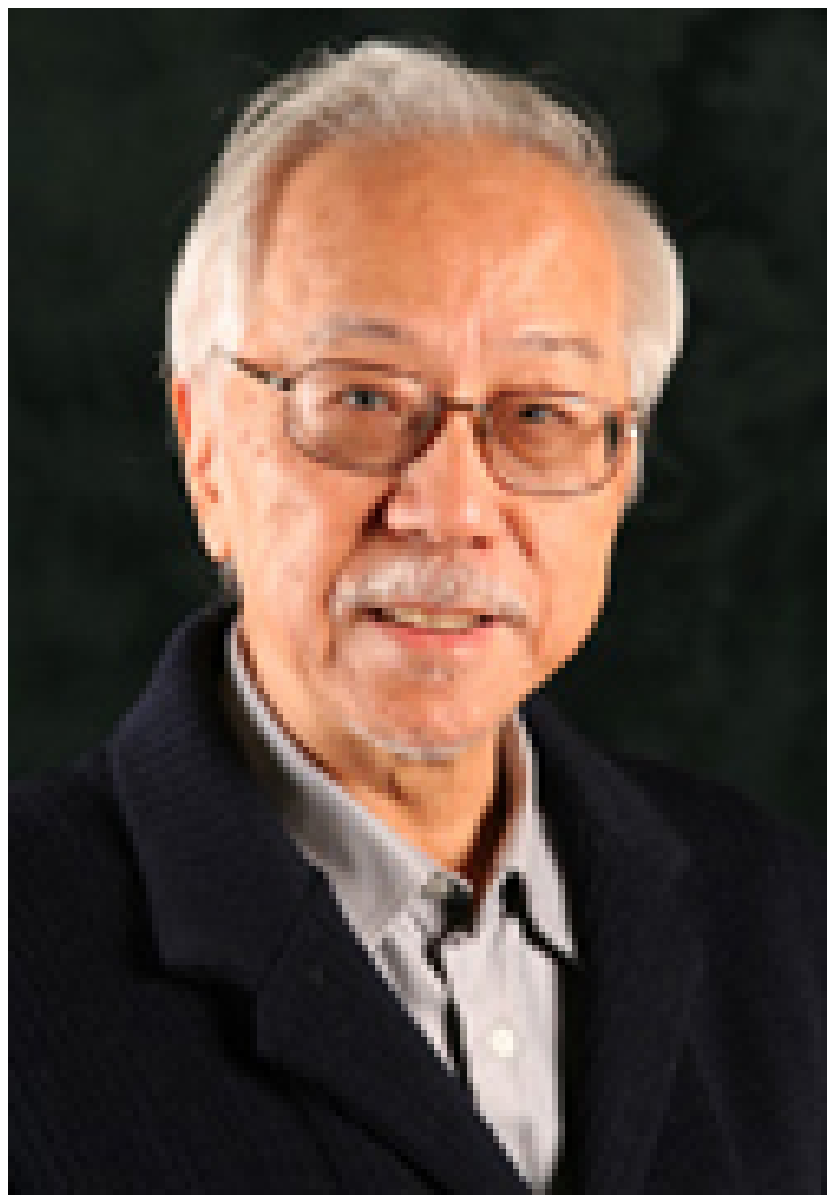}}]{Thomas S. Huang}
Prof. Thomas Huang received his Sc.D. from MIT in 1963. He is a full-time faculty with Beckman Institute at University of Illinois at Urbana-Champaign. He was William L. Everitt Distinguished Professor in the Department of Electrical and Computer Engineering and the Coordinated Science Lab (CSL), and was a full-time faculty member in the Beckman Institute Image Formation and Processing and Artificial Intelligence groups. His professional interests are computer vision, image compression and enhancement, pattern recognition, and multimodal signal processing.
\end{IEEEbiography}




\end{document}